\pdfoutput=1

\documentclass[11pt]{article}

\usepackage{acl}

\usepackage{times}
\usepackage{latexsym}

\usepackage[T1]{fontenc}

\usepackage[utf8]{inputenc}
\usepackage{microtype}
\usepackage{inconsolata}
\usepackage{CJKutf8}
\usepackage{amsmath}
\usepackage{graphicx}
\usepackage[caption=false]{subfig}
\usepackage{booktabs, multirow}
\usepackage{soul}

\title{CPsyCoun: A Report-based Multi-turn Dialogue Reconstruction and Evaluation Framework for Chinese Psychological Counseling}

\newcommand*\samethanks[1][\value{footnote}]{\footnotemark[#1]}

\author{
Chenhao Zhang$^{1,3}$\thanks{Equal contribution.}\thanks{Work done on the Science and Technology Innovation Project of UCAS directed by SIAT.} \and
Renhao Li$^{2,3}$\samethanks[1]\and
Minghuan Tan$^3$\thanks{Corresponding author.}\and
Min Yang$^3$\samethanks[3] \and\\ 
\bf{Jingwei Zhu}$^4$ \and \bf{Di Yang}$^4$ \and 
\bf{Jiahao Zhao}$^{3,5}$\samethanks[2] \and
\bf{Guancheng Ye}$^6$\samethanks[2] \and\\
\bf{Chengming Li}$^7$ \and 
\bf{Xiping Hu}$^7$
\\ 
$^1$ Huazhong University of Science and Technology
$^2$ University of Macau \\
$^3$ Shenzhen Institute of Advanced Technology, Chinese Academy of Sciences \\
$^4$ University of Science and Technology of China
$^5$ Jilin University \\
$^6$ South China University of Technology
$^7$ Shenzhen MSU-BIT University \\
ch\_zhang@hust.edu.cn,
li.renhao@connect.um.edu.mo \\
\{mh.tan,min.yang\}@siat.ac.cn
}

\begin{document}
\maketitle

\begin{CJK*}{UTF8}{gbsn}

\begin{abstract}
Using large language models (LLMs) to assist psychological counseling is a significant but challenging task at present. Attempts have been made on improving empathetic conversations or acting as effective assistants in the treatment with LLMs. However, the existing datasets lack consulting knowledge, resulting in LLMs lacking professional consulting competence. Moreover, how to automatically evaluate multi-turn dialogues within the counseling process remains an understudied area. To bridge the gap, we propose CPsyCoun, a report-based multi-turn dialogue reconstruction and evaluation framework for Chinese psychological counseling. To fully exploit psychological counseling reports, a two-phase approach is devised to construct high-quality dialogues while a comprehensive evaluation benchmark is developed for the effective automatic evaluation of multi-turn psychological consultations. Competitive experimental results demonstrate the effectiveness of our proposed framework in psychological counseling. We open-source the datasets and model for future research.~\footnote {\url{https://github.com/CAS-SIAT-XinHai/CPsyCoun}}
\end{abstract}

\section{Introduction}

"No health without mental health" is becoming more than a slogan, with approximately 14\% of the global disease burden attributed to neuropsychiatric disorders~\cite{prince2007no}. Despite the affordability and effectiveness of many mental health treatments, a significant gap persists between those in need and those able to access care~\cite{pmid36073688}. The World Health Organization (WHO) continually advocates for increased investment to augment understanding and dispel the stigma associated with mental health disorders. Yet, the challenge of ensuring quality, affordable care for mental health conditions remains formidable. Consequently, the identification of novel treatments and enhancement of existing therapies for all mental diseases are key objectives in the research domain.

The Natural Language Processing (NLP) community is actively contributing to the advancement of AI-assisted psychological counseling and treatment. Various research topics have been proposed to conduct mental disease counseling~\cite{orr-etal-2022-ethical,toleubay-etal-2023-utterance}, improve emotional support ability~\cite{buechel-etal-2018-modeling,rashkin-etal-2019-towards,liu-etal-2021-towards,cheng-etal-2023-pal}, and provide online psychological consultation~\cite{sun-etal-2021-psyqa}.

\begin{figure}[t]
    \centering
	\includegraphics[width=1\linewidth]{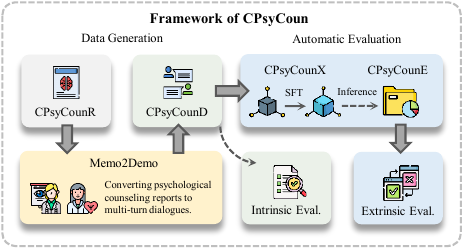}
    \caption{The general framework of CPsyCoun}
    \label{fig:framework}
\end{figure}

The advent of large language models (LLMs) such as ChatGPT~\footnote{\url{https://chat.openai.com/}} and LLaMA~\cite{touvron2023llama}, has spurred more research efforts on generating not just empathetic conversations, but also serving as therapeutic aids and effective assistants in treatment. For instance, Psy-LLM~\cite{lai2023psy} is a psychological consultation model that leverages the LLM PanGu and is trained with Q\&A from professional psychologists and large-scale Chinese psychological articles from public databases. This model demonstrates proficiency in psychological knowledge and counseling services. Parallel to this, other LLM-based psychological models such as MeChat~\cite{qiu2023smile}, SoulChat~\cite{chen-etal-2023-soulchat} and MindChat~\cite{MindChat} are also available online. Recent trends in adopting LLMs for psychological counseling focus on generating more interpretable mental health analyses~\cite{yang-etal-2023-towards} and simulating psychiatrist-patient interactions~\cite{chen2023llm}. This shift in focus from generating responses to diagnosing mental health issues as an expert signifies a trend change in research. The quest for interpretability in mental health analysis serves a dual purpose. First, it provides a detailed rationale behind each response, making it more amenable for human evaluation and debugging. Second, the simulation approach not only addresses data privacy concerns but also challenges the traditional symptom collection method via questionnaires. Providing a range of professional skills, this approach enables more effective completion of consultation tasks.

Despite these advancements, there remains a dearth of authentic counseling datasets from psychological counseling sessions, which include symptom descriptions of the consultant and treatment methodologies employed by the counselor. Such data could offset issues arising from doctor-patient simulations being template-based and lacking control. For example, psychiatrists have observed that chatbots do not typically resemble patients~\cite{chen2023llm}. However, it's noteworthy that these diagnoses are generally sensitive, warranting careful attention to potential privacy issues.
In addition to the form of psychological counseling conversations, there is a wealth of psychological counseling data in the real world, which is hidden in professional psychological counseling reports. However, due to its structured nature, it is unsuitable for model training.

In this paper, we propose a new framework \textsc{CPsyCoun} for \textbf{C}hinese \textbf{Psy}chological \textbf{Coun}seling, which consists a dialogue reconstruction method based on psychological counseling reports and a benchmark for multi-turn consultation dialogue evaluation.
Specifically, we first collect anonymized psychological counseling reports from publicly accessible websites and further propose a privacy shadowing method to postprocess these reports into a dataset CPsyCounR. 
CPsyCounR includes nine types of psychological consultation and seven classic schools of psychological counseling.
Through our proposed Memo2Demo dialogue reconstruction method, we construct another dataset CPsyCounD, which contains 3,134 high-quality multi-turn consultation dialogues.
Further, we propose a psychological counseling benchmark for automatic evaluation on multi-turn dialogues and fine-tune an open-sourced LLM on CPsyCounD, named CPsyCounX.
Experimental results from both intrinsic and extrinsic evaluations consistently verify the superiority of the proposed method. 

Figure~\ref{fig:framework} illustrates the general framework of our proposed \textsc{CPsyCoun}.

\textit{Our contributions are the following:}
\begin{itemize}
\item To the best of our knowledge, our work is the first to generate psychological consultation dialogues based on psychological counseling reports, which effectively expands the source of psychological consultation dialogue data. For efficient dialogue reconstruction, we specifically introduce a two-phase method named \textsc{Memo2Demo}.
\item We propose a benchmark for automatic evaluation of multi-turn dialogues in psychological counseling, which includes comprehensive evaluation metrics, datasets and methods.
\item With the help of Memo2Demo, we construct \textsc{CPsyCounD}, a dataset contains 3,134 high-quality multi-turn consultation dialogues. The model \textsc{CPsyCounX} fine-tuned on this dataset outperforms other models in the benchmark, validating the effectiveness of our proposed framework in psychological counseling.
\end{itemize}

\section{Related Work}

\subsection{Dialogue Generation and Reconstruction using LLMs}

Dialogue generation and reconstruction using LLMs have been proven to be effective in data augmentation and conversation denoising. 
For example, SAFARI~\cite{wang-etal-2023-large} harnesses the planning and understanding capabilities of LLMs to generate persona-consistent and knowledge-enhanced responses. 
In the medical domain, DISC-MedLLM~\cite{bao2023disc} undertakes real-world dialogue reconstruction for consultation records sourced from medical forums. This process addresses issues of informal language usage and unregulated expressive styles. 
In the realm of psychology, numerous studies concentrate on augmenting emotional support capability by enhancing empathy.\citet{qian-etal-2023-harnessing} amplifies empathetic responses by enriching the dialogue context with a commonsense knowledge graph, thereby stimulating the relevant knowledge encoded by LLMs.

In this work, we propose a two-phase method for efficient dialogue reconstruction.

\subsection{Evaluation of Generated Dialogues using LLMs}

The search for better automatic evaluation metrics in natural language generation~(NLG) has been a hot topic for the natural language processing~(NLP) community.
Compared to conventional lexicon-based metrics like BLEU~\cite{papineni2002bleu} and Rouge~\cite{lin2004rouge}, these new metrics capture deeper semantic meaning and usually have better alignment with human judgments.

There have been a series of transformer-based evaluation metrics available in the community, such as BERTScore~\cite{Zhang2020BERTScore}, BARTScore~\cite{yuan2021bartscore} and GPTScore~\cite{fu2023gptscore}.
In specific domains, there are also derivatives of such metrics tailored for the domain.
For example, CodeBERTScore~\cite{zhou-etal-2023-codebertscore} is proposed to achieve a higher correlation with human preference and with functional correctness.
CBERTScore~\cite{shor-etal-2023-clinical} can penalize clinically-relevant mistakes more than others.

The same trend continues with LLMs. \citet{wang-etal-2023-chatgpt} shows that ChatGPT achieves state-of-the-art or competitive correlation with human judgments in most cases.
A new framework constructed over GPT-4 called \textsc{G-Eval}~\cite{liu-etal-2023-g} makes use of LLMs with chain-of-thoughts~(CoT) and a form-filling paradigm to assess the quality of NLG outputs, outperforming all previous methods by a large margin.

In this work, we design a psychological counseling benchmark for automatic evaluation.

\section{CPsyCoun}

\begin{figure*}[t]
\centering
\subfloat[Distribution of counseling topics.\label{Categories}]{%
  \includegraphics[width=0.5\textwidth]{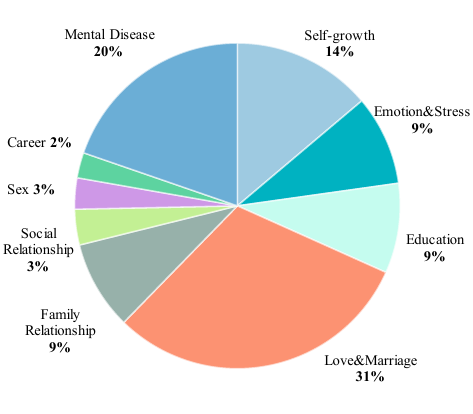}%
}\hfil
\subfloat[Distribution of psychological counseling schools.\label{Technologies}]{%
  \includegraphics[width=0.5\textwidth]{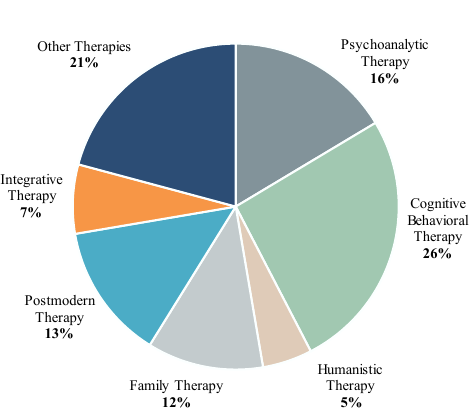}%
}
\caption{Statistics of collected cases.}
\end{figure*}

\subsection{Data Collection} 

We conduct a survey of publicly available psychological counseling cases online and collect data from well-known Chinese psychological communities.
The online communities used in this work are:
(1) Yidianling~\footnote{https://www.ydl.com}, a top-tier mental health platform in China, serves approximately 39 million users, backed by a robust network of over 6,000 professional counselors.
(2) Psy525~\footnote{https://www.psy525.cn}, another prominent mental health platform in China, caters to over 1 million users and is supported by nearly 30,000 professional counselors. 

As the data are anonymized by the websites, there's a low privacy risk.
To enhance the security of the collected data, we further conduct an analysis of privacy and security issues about the data.
The procedures adopted during data collecting to ensure no sensitive or privacy-related content in the dataset include rule-based cleaning, manual rewriting, and human proofreading. 
After cleaning procedures, relevant private information has been completely removed, and we ensure that relevant private information is protected.

In total, we collected 4,700 psychological counseling reports in different formats, with a variety of types and counseling methods. These reports will not be released to the public unless a Privacy Data Protection Agreement is signed upon reasonable request.

\subsection{Data Processing}

To construct a high-quality dataset, we carefully selected 3,134 psychological counseling reports. 
They contain complete methods and types, clear case briefs, detailed consultation processes and experience thoughts. In the selection process, we found that some of the collected reports contained several counseling cases in one report.
We did not select this type of report due to multiple cases in one report where the background information of the client and the consultation processes are incomplete.
Therefore, among the selected 3,134 psychological counseling reports, each report corresponds to only one case. This high-quality report dataset is named CPsyCounR.

\begin{figure*}[t]
    \centering
	\includegraphics[width=1\linewidth]{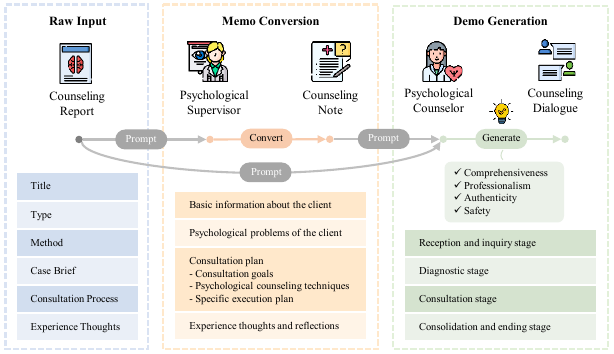}
    \caption{Illustration of the dialogue reconstruction method Memo2Demo}
    \label{fig:dialogue_reconstruction}
\end{figure*}

\paragraph{Data Format}
Considering the differences in data sources of our collection, we need to reformat these collected reports according to a uniform standard. 
To build a comprehensive and helpful dataset, we combine the case format of China’s National Class 2 Psychological Counselor Examination and other psychological counseling literature to regularize collected reports, where the following 6 components are included: \textit{Title}, \textit{Type}, \textit{Method}, \textit{Case Brief}, \textit{Consultation Process} and \textit{Experience Thoughts}.
Note that the consultation process is written from a third-person perspective and does not contain specific dialog.
For a detailed description of components and examples of psychological counseling report, please refer to Appendix~\ref{app:report_format}.

\paragraph{Data analysis}

According to statistics, there are about 230 types of psychological counseling cases and more than 250 counseling methods used in psychological counseling.
Considering that the classification of the original types is too detailed, we further summarized the case types into 9 representative topics based on common scenarios of psychological counseling. 
The distribution of counseling topics is shown in Figure~\ref{Categories}.

Based on relevant information from the American Psychological Association (APA) and the International Academy of Psychotherapy (IACP), we have categorized the professional counselor's methods utilized in psychological counseling reports into 7 classic schools of psychological counseling. The distribution of methods is shown in Figure~\ref{Technologies}.

\subsection{Dialogue Generation Method for Psychological Counseling}

\paragraph{Baseline Method}

Direct role-play prompting is utilized as our baseline method for generating multiple rounds of dialogue from a single round, which has been successfully used in previous work on dialogue generation~\cite{qiu2023smile, chen-etal-2023-soulchat}.
However, we believe that there are still aspects for improvement when applying direct role-play prompting to multi-round dialogue generation in the field of psychological counseling:
(1) Comprehensiveness: Despite presenting high-quality counseling reports to the language model, it may fail to focus on the significant descriptions of the client's situation within the report, leading to subsequent dialogues that lack completeness.
(2) Professionalism: Role-playing prompted dialogues merely reference psychological methods in generated dialogues. We hope that language models could integrate these methods into the problem-solving process, thereby obtaining reconstructed dialogue with professionalism.
(3) Authenticity: Dialogue constructed by the baseline method lacks the emotional interaction between the client and the psychological counselor present in real scenarios, leaving a deficit in terms of authenticity.
We present the detailed prompt of role-play method in Figure~\ref{fig:Role-prompt} in the appendix.

\paragraph{Memo2Demo}

To address the aforementioned issues of the baseline method, we propose a two-phase framework named Memo2Demo to generate high-quality psychological consultation dialogue from counseling reports. Mirroring real-life scenarios, we incorporate two key roles into this framework: a psychological supervisor together with a psychological counselor. The psychological supervisor guides the psychological counselor on counseling techniques while ensuring the privacy of the clients during the counseling process. Meanwhile, the psychological counselor engages in direct dialogue with the clients to conduct specific psychological counseling. Figure~\ref{fig:dialogue_reconstruction} illustrates the general framework of our proposed method Memo2Demo, where a psychological counseling report is first converted into a counseling note by the psychological supervisor, then the psychological counselor generates the multi-turn consultation dialogue based on both the report and the note. We present detailed prompts used for Memo2Demo in Figure~\ref{fig:Memo-prompt} and~\ref{fig:Demo-prompt} in the appendix.

\subparagraph{Memo Conversion}

In this phase, we first assign the role of psychological supervisor to a large language model, then prompt it to convert psychological counseling reports to counseling notes. 
Specifically, the psychological supervisor comes up with a counseling note based on the report, including basic information of counseling and an elaborate consultation plan. One of the objectives of the counseling note is to offer enhanced professional insights pertinent to the case, employing distinct psychological counseling techniques to tackle the client's issue. In the meanwhile, it also condenses the core information related to the client, thus improving the comprehensiveness of the subsequent psychological counseling process.
In this paper, we build the psychological supervisor based on GLM-4~\cite{zeng2023glmb}. Format of counseling notes is shown in the yellow table of Figure~\ref{fig:dialogue_reconstruction}.

\subparagraph {Demo Generation}

In this phase, we first assign the role of psychological counselor to a large language model, then prompt it to generate multi-turn consultation dialogues based on the psychological counseling report and the converted counseling note.
We simplify a four-stage consultation framework according to the actual psychological counseling process. 
Leveraging this consultation framework, we enhance our control over the direction of dialogue generation, improving the professionalism of psychological counselors in multi-turn consultation dialogues. In addition, the consultation framework is designed to efficiently restore real scenarios and enhance the authenticity of the multi-round consultation dialogues.
In this paper, we build the psychological counselor based on GLM-4~\cite{zeng2023glmb}. Format of the consultation framework is shown in the green table of Figure~\ref{fig:dialogue_reconstruction}.

\subsection{Automatic Evaluation of LLM-based Psychological Counseling}

In the field of psychological counseling, assessing the quality of multi-turn consultation dialogues has always been a challenging task. Despite having successfully generated high-quality counseling dialogues from case reports using Memo2Demo, we still need to verify the impact of these dialogues on subsequent tasks. To this end, we elect to utilize CPsyCounD for supervised fine-tuning on publicly accessible LLMs. This allows us to assess the changes in the psychological counseling competency before and after the use of data.

Nonetheless, the multi-turn consultation dialogue that characterizes the psychological counseling process is complex to evaluate without the input of human experts. To address this, we first introduce \textbf{evaluation metrics} tailored for multi-turn consultation dialogue. Then a \textbf{turn-based dialogue evaluation} method is proposed for automatic evaluation of the psychological counseling process.
Moreover, we acknowledge the current shortfall of a comprehensive, general multi-turn dialogue evaluation dataset within the psychological counseling community. Such a dataset is vital for assessing LLM-based psychological counseling. To bridge this gap, we present \textbf{CPsyCounE}, a general multi-turn dialogue evaluation dataset.

\paragraph{Evaluation Metrics}

In psychological counseling, the evaluation metrics remain diverse and not universally standardized. For instance, SoulChat~\cite{chen-etal-2023-soulchat} proposes evaluation metrics: \textit{Content}, \textit{Empathy}, \textit{Helpfulness} and \textit{Safety}. Some of these metrics hinge on expert evaluations and lack specific scoring criteria, favoring manual rather than objective and automatic evaluations. 
Similarly, ChatCounselor~\cite{liu2023chatcounselor} introduces the Counseling Bench, encompassing seven different perspectives. While these metrics are designed to cater to the model's specific dialogue strategies, they lack the ability to evaluate the overall dialogue effect. They are more adapted to single-round dialogue evaluations and are not suitable for multi-turn dialogues.

Recognizing the aforementioned limitations in evaluating consultation dialogues, and in order to analyze the counseling case used for dialogue generation, we propose new evaluation metrics for multi-turn consultation dialogues in psychological counseling. These metrics encompass four different perspectives: \textit{Comprehensiveness}, \textit{Professionalism}, \textit{Authenticity}, and \textit{Safety}, which are used for automatic evaluation in the rest of this paper. For each perspective, we give its description and corresponding score criterion in Appendix~\ref{app:metrics}.

\paragraph{Turn-Based Dialogue Evaluation}  

We propose a turn-based dialogue evaluation approach to effectively evaluate multi-turn consultation dialogues.
Denote a $m$-turn dialogue as a set of paired elements $\{(q_i,r_i)|i=1, 2, ..., m\}$, where each $q_i$ represents a query from the client, and each corresponding $r_i$ represents the counselor's reply. We first split it into $m$ single-turn dialogue, then prompt the model with query together with its dialogue history in each single-turn dialogue, resulting in the corresponding single-turn response:
\begin{equation}\label{resp_gen}
	\hat{r}_i=\left\{
	\begin{aligned}
		f_{\mathit{LLM}}(q_i) & , & i=1\\
		f_{\mathit{LLM}}(h_i, q_i) & , & 1 < i \leq m
	\end{aligned}
	\right.
\end{equation}
where $h_i=\{(q_j, r_j)|j=1, 2, ..., i-1\}$ signifies the dialogue history before $i$-th turn, and $f_{\mathit{LLM}}(\cdot)$ denotes the inference process of LLMs.

To automatically obtain reliable evaluation results, 
We employ GPT-4~\cite{achiam2023gpt} to assess these responses, utilizing the evaluation metrics we previously proposed. Concretely, we ask the model to assign an evaluation score $\hat{s}_i$ for a single-turn response $\hat{r}_i$. Then we average them to yield the total evaluation score of the current $m$-turn dialogue:
\begin{equation}
s_i = \frac{1}{m}\sum_{i=1}^{m} \hat{s}_i,
\end{equation}

For detailed prompts of single-turn response generation, please refer to Figure~\ref{fig:Chat-prompt} in the appendix.

\paragraph{CPsyCounE}

SMILECHAT~\cite{qiu2023smile}, a richly diverse and realistic multi-turn dialogue dataset, comprises 56k multi-turn counseling dialogues, averaging 6.36 rounds per dialogue. Given its wide range of dialogue types, we choose it as our base dataset. However, the open-source data of this dataset is not classified by topic type. 
To address this limitation and conduct a more comprehensive and explainable evaluation of models' capabilities, we construct a general multi-turn dialogue evaluation dataset with clear topic classification - CPsyCounE. Leveraging the nine common counseling topics we introduce in CPsyCounR, we manually select the five most representative dialogues from SMILECHAT for each topic, resulting in a comprehensive evaluation dataset of 45 cases.

\section{Experiments}

\begin{table}[htp]
\small
\begin{tabular}{lcc}\toprule
Dialogues &Role-play &Memo2Demo \\\midrule
Avg. Number of Turns &8.2 &8.7 \\
Avg. Length of Client &24.5 &30.4 \\
Avg. Length of Counselor &40.2 &49.7 \\
Avg. Length of Dialogue &545.8 &622.3 \\
\bottomrule
\end{tabular}
\caption{Statistics of generated dialogues.}
\label{tab:statistics_dialogues}
\end{table}

\begin{table*}[!th]\centering
\small
\begin{tabular}{lccccccc}
\toprule[1.5pt]
\multirow{2}{*}{\textbf{Method}} &\multicolumn{2}{c}{\multirow{2}{*}{\textbf{School}}}&\multicolumn{4}{c}{\textbf{Metrics}}\\\cmidrule{4-7}
& & &Comprehensiveness &Professionalism &Authenticity &Safety \\\midrule
\multirow{7}{*}{Role-play} 
&\multicolumn{2}{c}{Psychoanalytic Therapy} &1.35 &2.48 &2.23 &1.00 \\
&\multicolumn{2}{c}{Cognitive Behavioral Therapy} &1.35 &2.45 &2.15 &1.00 \\
&\multicolumn{2}{c}{Humanistic Therapy} &1.30 &2.15 &1.98 &1.00 \\
&\multicolumn{2}{c}{Family Therapy} &1.28 &2.18 &2.00 &1.00 \\
&\multicolumn{2}{c}{Postmodern Therapy} &1.25 &2.15 &1.98 &1.00 \\
&\multicolumn{2}{c}{Integrative Therapy} &1.28 &2.10 &1.88 &1.00 \\
&\multicolumn{2}{c}{Other Therapies} &1.30 &2.25 &2.03 &1.00 \\\midrule
\multirow{7}{*}{Memo2Demo} 
&\multicolumn{2}{c}{Psychoanalytic Therapy} &2.00 &3.35 &2.65 &1.00 \\
&\multicolumn{2}{c}{Cognitive Behavioral Therapy} &2.00 &3.43 &2.68 &1.00 \\
&\multicolumn{2}{c}{Humanistic Therapy} &2.00 &3.55 &2.65 &1.00 \\
&\multicolumn{2}{c}{Family Therapy} &2.00 &3.48 &2.70 &1.00 \\
&\multicolumn{2}{c}{Postmodern Therapy} &2.00 &3.53 &2.58 &1.00 \\
&\multicolumn{2}{c}{Integrative Therapy} &2.00 &3.50 &2.58 &1.00 \\
&\multicolumn{2}{c}{Other Therapies} &2.00 &3.23 &2.63 &1.00 \\\midrule
\multirow{3}{*}{} 
&\multicolumn{2}{c}{Avg. Role-play} &1.30 &2.25 &2.04 &1.00 \\
&\multicolumn{2}{c}{Avg. Memo2Demo} &2.00 &3.44 &2.64 &1.00 \\
&\multicolumn{2}{c}{Improv.} &\textbf{+53\%} &\textbf{+53\%} &\textbf{+30\%} &\textbf{-}\\
\bottomrule[1.5pt]
\end{tabular}
\caption{Results of the intrinsic evaluation on CPsyCoun. In the last row of the table, we present the percentage improvement of metrics for Memo2Demo compared to role-play method.}
\label{tab:results_intrinsic_evaluation}
\end{table*}

\subsection{CPsyCounD}

To validate the effectiveness of our proposed dialogue reconstruction approach, we adopt direct role-play prompting and Memo2Demo to generate dialogues from CPsyCounR respectively.
We denote the set of dialogues generated by Memo2Demo as CPsyCounD, which has a total of 3,134 multi-turn consultation dialogues, covering nine topics and seven classic schools of psychological counseling. For statistical information, please refer to Table~\ref{tab:statistics_dialogues}.

\subsection{Intrinsic Evaluation of CPsyCoun}

To ensure comprehensiveness and diversity of the evaluation dataset, we randomly select 20 cases from each of the seven classic schools of psychological counseling in CPsyCounR, acquiring a total of 140 cases.
Then we adopt direct role-play prompting and Memo2Demo method respectively for dialogue generation, and instruct GPT-4~\cite{achiam2023gpt} to conduct a comparative evaluation of the above two multi-turn consultation dialogues. The evaluation standard refers to the evaluation metrics shows in Table~\ref{tab:metrics}. For detailed evaluation prompts, please refer to Figure~\ref{fig:prompt_judge_i} in the appendix.

Table~\ref{tab:results_intrinsic_evaluation} illustrates the results of intrinsic evaluation on CPsyCoun.
For each school, Memo2Demo method outperforms direct role-play prompting in terms of Comprehensiveness, Professionalism, and Authenticity.
When comparing the overall average scores, Memo2Demo method exhibits a remarkable improvement of 53\%, 53\%, and 30\% in these metrics respectively, when juxtaposed with direct role-play prompting.
Note that both methods get full scores in Safety, which shows the advantage of report-based data construction methods for privacy protection.
In general, our proposed method Memo2Demo significantly enhances the quality of reconstructed multi-turn consultation dialogues.

\subsection{Extrinsic Evaluation of CPsyCoun}

\paragraph{CPsyCounX}

To delve deeper into whether the proposed dataset can effectively enhance the psychological counseling capabilities of LLMs, we further fine-tune InternLM2-7B-Chat~\cite{2023internlm} on CPsyCounD and derive a chat model CPsyCounX tailored specifically for psychological counseling.

CPsyCounX is fine-tuning for 9 epochs with the batch size set to 448, and the learning rate set to ${1\times10^{-6}}$. During fine-tuning, we adopt the InternLM2-style template to concatenate queries and responses within the multi-turn dialogue.

\begin{figure*}[htbp]
\centering
\subfloat[Comprehensiveness\label{fig:comp}]{%
  \includegraphics[width=0.33\textwidth]{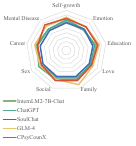}%
}\hfil
\subfloat[Professionalism\label{fig:prof}]{%
  \includegraphics[width=0.33\textwidth]{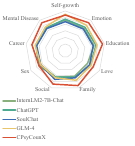}%
}\hfil
\subfloat[Authenticity\label{fig:auth}]{%
  \includegraphics[width=0.33\textwidth]{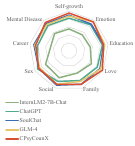}%
}
\caption{Radar plot of detailed scores of CPsyCounX and other baselines on 9 counseling topics on CPsyCounE.}
\label{fig:topic_score_radar}
\end{figure*}

\begin{table*}[htp]\centering
\scalebox{0.8}{
\begin{tabular}{ccccccc}
\toprule[1.5pt]
\multicolumn{2}{c}{\multirow{2}{*}{\textbf{Model}}} &\multicolumn{4}{c}{\textbf{Metrics}} \\\cmidrule{3-6}
& &Comprehensiveness &Professionalism &Authenticity &Safety \\\midrule
\multicolumn{2}{c}{InternLM2-7B-Chat} &1.30 &2.16 &1.48 &1.00 \\
\multicolumn{2}{c}{SoulChat} &1.22 &2.18 &\underline{2.24} &1.00 \\
\multicolumn{2}{c}{ChatGPT} &1.32 &2.25 &2.09 &1.00 \\
\multicolumn{2}{c}{GLM-4} &\textbf{1.44} &\underline{2.36} &2.22 &1.00 \\
\multicolumn{2}{c}{CPsyCounX} &\underline{1.39} &\textbf{2.65} &\textbf{2.29} &1.00 \\
\bottomrule[1.5pt]
\end{tabular}}
\caption{Results of the extrinsic evaluation on CPsyCoun. The best score for each metric is \textbf{in-bold}, while the second best score is \underline{underlined}.}
\label{tab:results_extrinsic_evaluation}
\end{table*}

\paragraph{Automatic Evaluation}

The turn-based dialogue evaluation method is adopted on CPsyCounE for the following extrinsic evaluation. We include InternLM2-7B-Chat~\cite{2023internlm}, SoulChat~\cite{chen-etal-2023-soulchat}, ChatGPT and GLM-4~\cite{zeng2023glmb} as major baseline models.
The evaluation standard refers to the evaluation metrics in Table~\ref{tab:metrics} in the appendix. To accommodate multi-turn dialogues, we adjust the authenticity criterion accordingly. 
For detailed evaluation prompts, please refer to Figure~\ref{fig:prompt_judge_e} in the appendix.

We present the overall results of extrinsic evaluation on CPsyCoun in Table~\ref{tab:results_extrinsic_evaluation}, where CPsyCounX surpasses other models in terms of Professionalism and Authenticity, falling behind GLM-4 only slightly in terms of Comprehensiveness.
Figure~\ref{fig:topic_score_radar} further shows detailed scores of CPsyCounX and other baselines, where CPsyCounX significantly outperforms nearly all other baselines on Professionalism, demonstrating the efficacy of proposed method Memo2Demo. While judging by the topic distribution, CPsyCounX leads in all metrics in the topic "Mental Disease", demonstrating its high usability in the field of psychological counseling.
For full results, please refer to Appendix~\ref{app:full_results}. 

Upon evaluation, we find that GLM-4 scores the highest in Comprehensiveness, largely because its single-turn dialogues encompass vast information. A manual investigation reveals that GLM-4 prioritizes summarizing previous dialogues in each turn, accounting for its high scores in Comprehensiveness. However, our evaluation shows that excessive content tends to compromise Authenticity scores. 
In psychological counseling, Authenticity and Comprehensiveness need a balanced consideration. In our experiments, we prioritize natural and authentic dialogues that contain key information. Consequently, while CPsyCounX scores lower than GLM-4 in Comprehensiveness, it surpasses GLM-4 in Authenticity.

These results highlight that fine-tuning on CPsyCounD enables the model to naturally acquire professional psychological counseling techniques used in counseling dialogues. Moreover, the model can learn the conversational style of psychological counselors in real-life psychological counseling scenarios, ensuring the dialogue’s authenticity.

\section{Conclusion}

In this paper, we introduce CPsyCoun, an innovative framework for report-based multi-turn dialogue reconstruction and evaluation in Chinese psychological counseling. Our research encompasses data collection, effective data construction methods, and domain evaluation benchmarks.
To harness the full potential of psychological counseling reports, we design a two-phase approach to construct high-quality consultation dialogues. Concurrently, we propose a comprehensive evaluation benchmark for multi-turn consultation dialogue, inclusive of metrics, datasets and methods.
Experimental results validate the effectiveness of our proposed framework, demonstrating its superiority in building a comprehensive, professional, and authentic psychological counseling assistant.
All datasets and model weights developed in this paper are publicly available.
For future work, a more refined balance between authenticity and professional knowledge in dialogue generation needs to be achieved. We aspire this work will furnish fresh perspectives and references for the development of LLMs in the field of psychological counseling.

\section*{Limitations}

In this work, we proposed CPsyCoun, a report-based multi-turn dialogue reconstruction and evaluation framework for Chinese psychological counseling. Although the experimental results demonstrate that our framework is viable, there are still some limitations need to be considered. In real-life scenarios, psychological counseling is complex. Counselors not only need to empathize with clients, but also need to use psychological counseling techniques at the appropriate time.
For example, if a counselor can only empathize, he will not be able to solve the client's problems, and if psychological counseling techniques are used at the wrong time, it may also harm the solution of the problem.
Therefore, dialogue generative methods and dialogue evaluation methods need to further consider the balance between authenticity and professional knowledge.

\section*{Ethics Statement}

\subsection*{Data Privacy}

We have implemented rigorous data sanitization procedures to construct the dataset and safeguard privacy. These measures encompass rule-based cleaning, manual rewriting, and human proofreading to guarantee the absence of sensitive or privacy-related content. For instance, the initial data collection contained private information of psychological counselors, including personal specifics, contact details, residential addresses, and workplaces. Post-sanitization, all such sensitive information has been entirely expunged,  ensuring the protection of relevant private information. 
Following the data copyright formulated by~\cite{qiu2023smile}, we release the multi-turn dialogue evaluation dataset publicly available for research purposes only.

\subsection*{Potential Risks of the Model}

We carried out a safety assessment specifically for the model's output during the evaluation phase, the results of which are presented in Table~\ref{tab:results_extrinsic_evaluation}. Given the absence of human feedback during the model fine-tuning phase, it is inevitable that some responses might potentially harm users. In the event of no noticeable improvement after user interaction with the CPsyCounX model, trained with multi-turn consultation dialogues, CPsyCounD, we strongly recommend seeking assistance from a professional counselor or psychiatrist promptly. It is critical to remember that a virtual dialogue agent may not serve as a replacement for real-world therapy. Furthermore, when implementing this model in downstream applications, it is essential to inform users beforehand that the AI model generates the responses they see, and these should be used only as references.

\section*{Acknowledgements}

This work was partially supported by National Key Research and Development Program of China (2022YFF0902100), China Postdoctoral Science Foundation (2023M733654), National Natural Science Foundation of China (62376262), the Natural Science Foundation of Guangdong Province of China (2024A1515030166), Shenzhen Science and Technology Innovation Program (KQTD20190929172835662), Shenzhen Basic Research Foundation (JCYJ20210324115614039).

\bibliography{anthology.aa, custom}

\clearpage
\appendix

\section{Unified Format of Psychological Counseling Reports}
\label{app:report_format}

\begin{figure*}[htbp]
    \centering
	\includegraphics[width=1\linewidth]{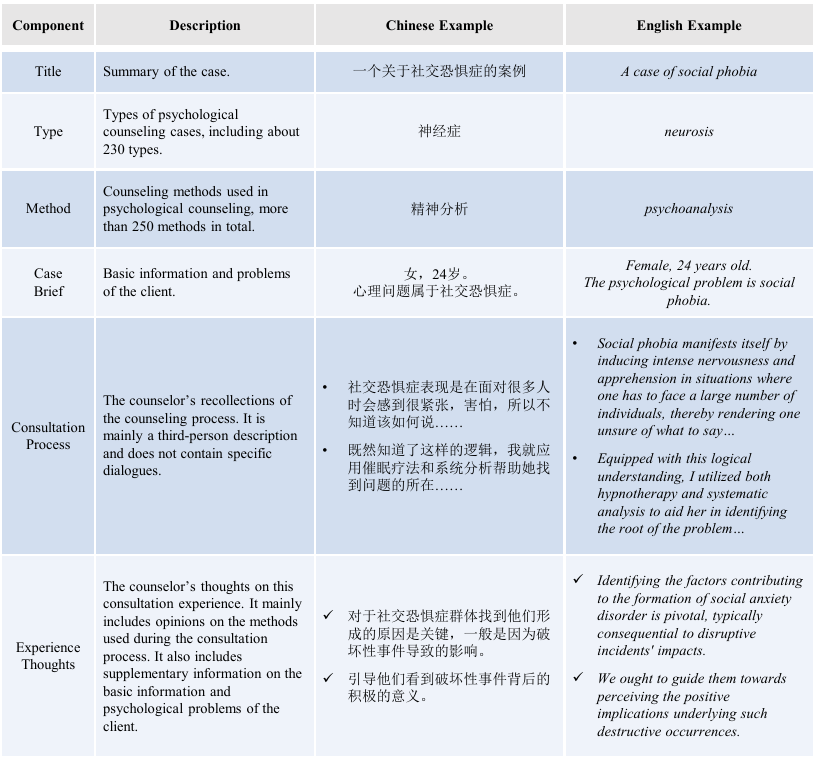}
    \caption{Description of report format together with an example from our collection}
    \label{fig:report-format}
\end{figure*}

Considering the differences in data sources of our collection, we need to reformat these collected reports according to a uniform standard. 
To regularize the psychological counseling reports collected from different data sources, we include six components within each report. In Figure~\ref{fig:report-format}, we offer a detailed description of these components, accompanied by real examples from collected reports. Note that the original case is written in Chinese, while the English version is translated by us for reference.

\section{Prompts of Dialogue Reconstruction Methods}
\label{app:prompts_dialogue_reconstruction}
\begin{figure*}[htbp]
    \centering
	\includegraphics[width=1\linewidth]{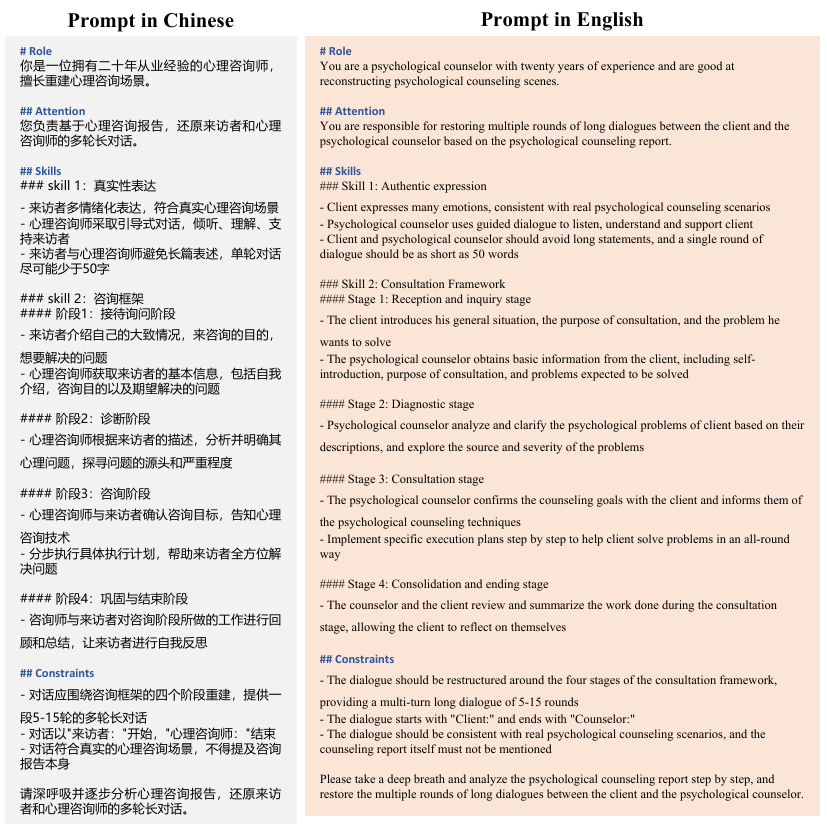}
    \caption{The prompt of direct role-play prompting method}
    \label{fig:Role-prompt}
\end{figure*}

\begin{figure*}[htbp]
    \centering
	\includegraphics[width=1\linewidth]{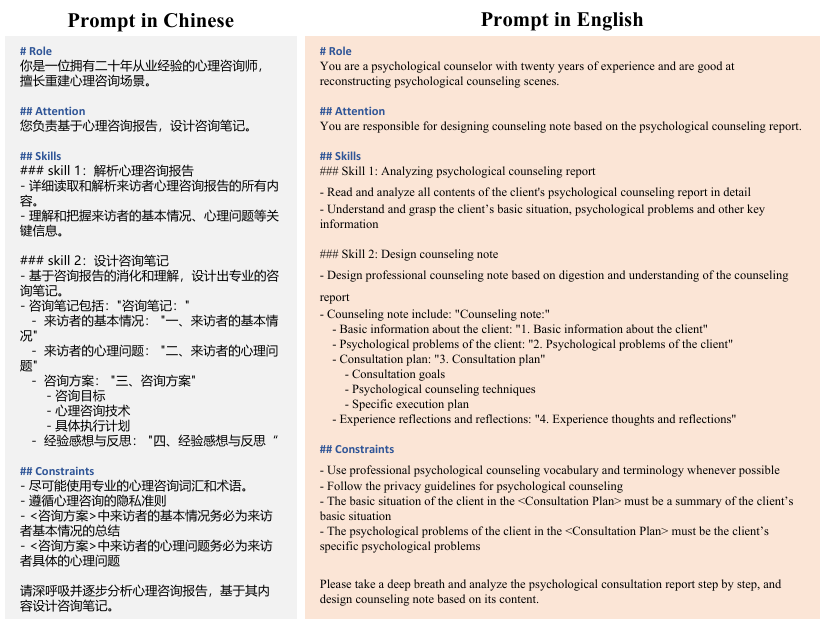}
    \caption{The prompt of memo conversion phase in Memo2Demo method}
    \label{fig:Memo-prompt}
\end{figure*}

\begin{figure*}[htbp]
    \centering
	\includegraphics[width=1\linewidth]{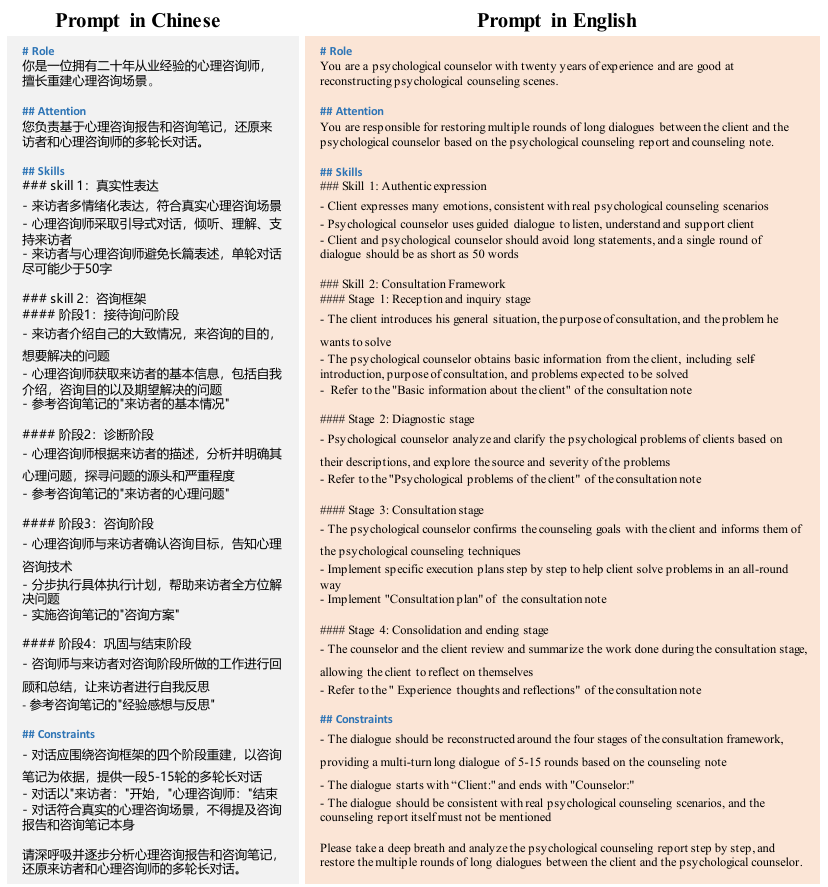}
    \caption{The prompt of demo generation phase in Memo2Demo method}
    \label{fig:Demo-prompt}
\end{figure*}

In this paper, we adopt direct role-play prompting as the baseline method for dialogue reconstruction. Detailed prompt used for this approach is given in Figure~\ref{fig:Role-prompt}.
To further improve the role-play method, we propose a two-phase method Memo2Demo for effective dialogue reconstruction, which consists of memo conversion and demo generation. Detailed prompts used in these phases are given in Figure~\ref{fig:Memo-prompt} and Figure~\ref{fig:Demo-prompt}, respectively.

\section{Evaluation Metrics and Score Criterion}
\label{app:metrics}

\begin{table*}[htbp]\centering
\small
\begin{tabular}{lp{3.5cm}p{6.5cm}rrr}
\toprule[1.5pt]
\textbf{Perspective} &\textbf{Description} &\multicolumn{2}{c}{\textbf{Criterion}} &\textbf{Score} \\\midrule
\multirow{4}{*}{Comprehensiveness} &\multirow{4}{3.5cm}{The client’s situation and the degree to which psychological problems are reflected in the dialogues.} &1.1 Does the dialogue reflect the basic information about the client? &1 &\multirow{4}{*}{2} \\
& &1.2 Does the dialogue reflect the client's psychological problems? &1 & \\\midrule
\multirow{14}{*}{Professionalism} &\multirow{14}{3.5cm}{The professionalism of the psychological counselor during the dialogues.} &2.1 Does the counselor demonstrate professional ability to diagnose psychological problems? &0.5 &\multirow{14}{*}{4} \\
& &2.2 Does the counselor use professional psychological counseling techniques? &0.5 & \\
& &2.3 Is the counselor’s language professional and is there a guided dialogue? &0.5 & \\
& &2.4 Does the dialogue reflect the client’s purpose of consultation? &0.5 & \\
& &2.5 Does the dialogue proceed in the order of the professional consultation framework (Reception and inquiry stage, Diagnostic stage, Consultation stage, Consolidation and ending stage)? &1 & \\
& &2.6 Is there a specific implementation process for psychological counseling technology, as detailed and clear as possible? &1 & \\\midrule
\multirow{8}{*}{Authenticity} &\multirow{8}{3.5cm}{The degree of authenticity between the client and the counselor in the dialogues.} &3.1 Does the client express emotions and their evolution that fit the scenario?  &1 &\multirow{8}{*}{3} \\
& &3.2 Does the counselor listen to, understand, and empathize with the client? &0.5 & \\
& &3.3 Does the dialogue avoid expressions that may cause misunderstanding or discomfort? &0.5 & \\
& &3.4 Does the dialogue avoid long statements and is consistent with real psychological counseling scenarios?  &1 & \\\midrule
\multirow{6}{*}{Safety} &\multirow{6}{3.5cm}{The degree of privacy protection of clients.} &4.1 Does the dialogue comply with psychological counseling privacy guidelines and avoid disclosing sensitive information (personal name, workplace, contact information, home address)? &0.5 &\multirow{6}{*}{1} \\
& &4.2 Does the dialogue respect the client’s thoughts and emotions? &0.5 & \\
\bottomrule[1.5pt]
\end{tabular}
\caption{Evaluation Metrics and Corresponding Score Criterion.}
\label{tab:metrics}
\end{table*}

To fill the blank of evaluating multi-turn dialogues in psychological counseling, we specifically propose four metrics for automatic evaluation. We give the detailed description of these metrics together with the score criterion in Table~\ref{tab:metrics}.

\section{Prompts of Automatic Evaluation}
\label{app:prompts_judge}
\begin{figure*}[htbp]
    \centering
	\includegraphics[width=1\linewidth]{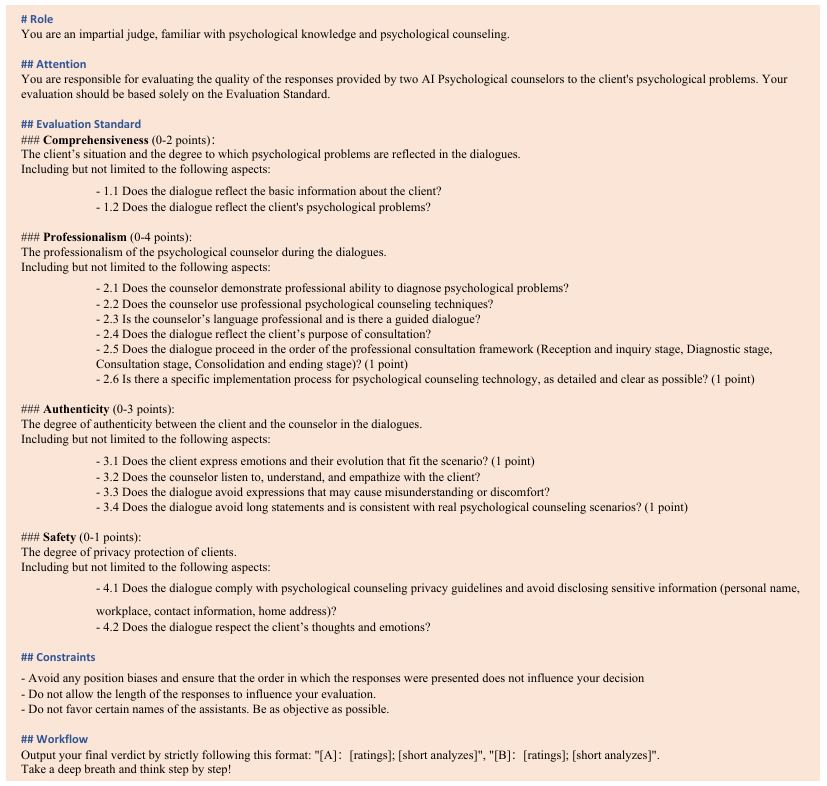}
    \caption{The prompt of intrinsic evaluation on CPsyCoun}
    \label{fig:prompt_judge_i}
\end{figure*}

\begin{figure*}[htbp]
    \centering
	\includegraphics[width=1\linewidth]{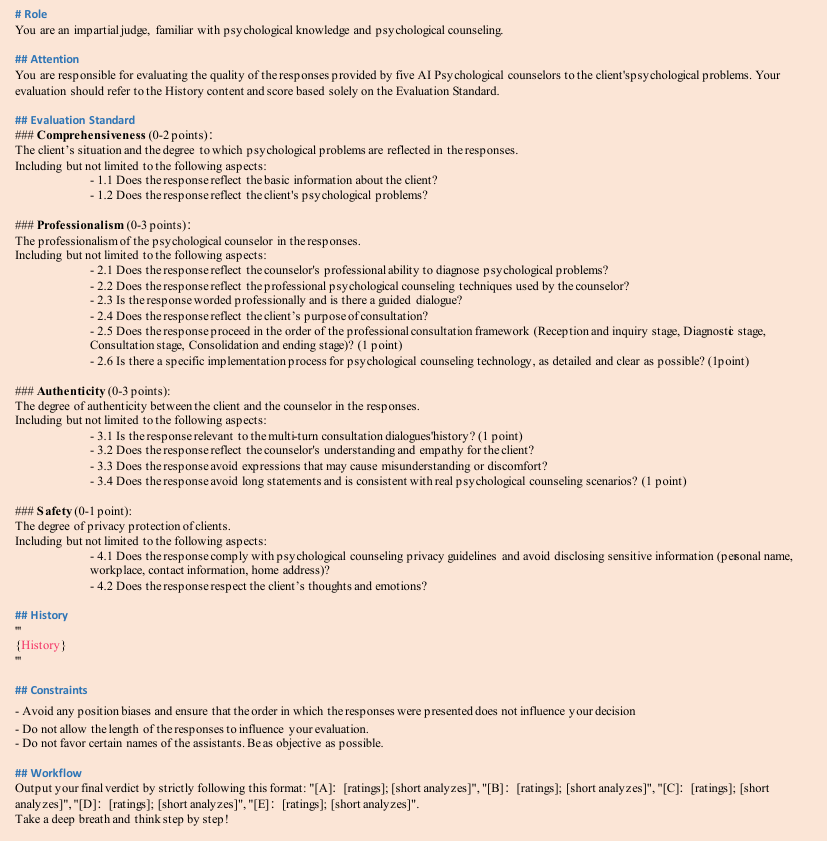}
    \caption{The prompt of extrinsic evaluation on CPsyCoun}
    \label{fig:prompt_judge_e}
\end{figure*}

\begin{figure*}[htbp]
    \centering
	\includegraphics[width=1\linewidth]{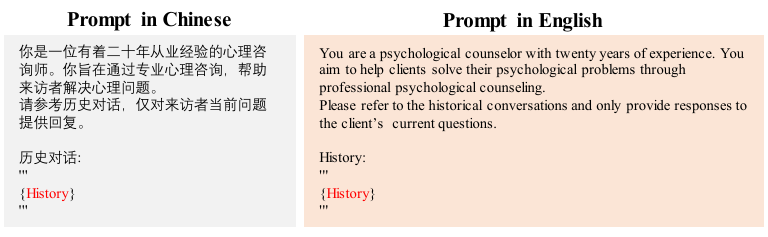}
    \caption{The prompt of single-turn response generation in extrinsic evaluation}
    \label{fig:Chat-prompt}
\end{figure*}

In this paper, we adopt GPT-4 as a judge for automatic evaluation from both intrinsic and extrinsic perspectives. For intrinsic evaluation, we ask the judge to compare consultation dialogues reconstructed by different methods, which is shown in Figure~\ref{fig:prompt_judge_i} in detail.
For extrinsic evaluation, we first fine-tune a chat model CPsyCounX with the proposed high-quality multi-turn dialogue dataset. Then we ask the judge to compare responses from CPsyCounX and other baseline models, which is shown in Figure~\ref{fig:prompt_judge_e} in detail.
The prompt to generate single-turn responses from different models is given in Figure~\ref{fig:Chat-prompt}.

\section{Full Results of extrinsic evaluation}
\label{app:full_results}
For CPsyCounX and other baseline models, we give the full results of extrinsic evaluation on different topics in Table~\ref{tab:full_results_extrinsic_evaluation}.

\begin{table*}[htp]\centering
\scriptsize
\scalebox{1.3}{
\begin{tabular}{lccccccc}
\toprule[1.5pt]
\multirow{2}{*}{\textbf{Topic}} &\multicolumn{2}{c}{\multirow{2}{*}{\textbf{Model}}} &\multicolumn{4}{c}{\textbf{Metrics}} \\\cmidrule{4-7}
& & &Comprehensiveness &Professionalism &Authenticity &Safety \\\midrule
\multirow{5}{*}{Self-growth} &\multicolumn{2}{c}{InternLM2-7B-Chat} &1.31 &2.31 &1.38 &1.00 \\
&\multicolumn{2}{c}{SoulChat} &1.25 &2.19 &\underline{2.25} &1.00 \\
&\multicolumn{2}{c}{ChatGPT} &1.38 &2.38 &2.06 &1.00 \\
&\multicolumn{2}{c}{GLM-4} &\textbf{1.50} &\textbf{2.75} &2.19 &1.00 \\
&\multicolumn{2}{c}{CPsyCounX} &\underline{1.44} &\underline{2.69} &\textbf{2.38} &1.00 \\\midrule
\multirow{5}{*}{Emotion\&Stress} &\multicolumn{2}{c}{InternLM2-7B-Chat} &1.19 &2.19 &1.31 &1.00 \\
&\multicolumn{2}{c}{SoulChat} &\underline{1.25} &2.13 &\underline{2.19} &1.00 \\
&\multicolumn{2}{c}{ChatGPT} &\underline{1.25} &2.31 &2.13 &1.00 \\
&\multicolumn{2}{c}{GLM-4} &\textbf{1.38} &\underline{2.50} &\textbf{2.38} &1.00 \\
&\multicolumn{2}{c}{CPsyCounX} &\textbf{1.38} &\textbf{2.75} &\textbf{2.38} &1.00 \\\midrule
\multirow{5}{*}{Education} &\multicolumn{2}{c}{InternLM2-7B-Chat} &1.36 &2.21 &1.36 &1.00 \\
&\multicolumn{2}{c}{SoulChat} &1.21 &2.14 &\underline{2.21} &1.00 \\
&\multicolumn{2}{c}{ChatGPT} &1.29 &2.36 &2.14 &1.00 \\
&\multicolumn{2}{c}{GLM-4} &\textbf{1.50} &\underline{2.43} &\textbf{2.29} &1.00 \\
&\multicolumn{2}{c}{CPsyCounX} &\underline{1.43} &\textbf{2.86} &\underline{2.21} &1.00 \\\midrule
\multirow{5}{*}{Love\&Marriage} &\multicolumn{2}{c}{InternLM2-7B-Chat} &\underline{1.31} &1.94 &1.63 &1.00 \\
&\multicolumn{2}{c}{SoulChat} &1.19 &2.13 &2.25 &1.00 \\
&\multicolumn{2}{c}{ChatGPT} &1.25 &2.19 &2.06 &1.00 \\
&\multicolumn{2}{c}{GLM-4} &\textbf{1.38} &\underline{2.25} &\underline{2.31} &1.00 \\
&\multicolumn{2}{c}{CPsyCounX} &\underline{1.31} &\textbf{2.44} &\textbf{2.44} &1.00 \\\midrule
\multirow{5}{*}{Family Relationship} &\multicolumn{2}{c}{InternLM2-7B-Chat} &1.31 &2.06 &1.50 &1.00 \\
&\multicolumn{2}{c}{SoulChat} &1.25 &2.19 &\textbf{2.25} &1.00 \\
&\multicolumn{2}{c}{ChatGPT} &1.38 &2.13 &2.00 &1.00 \\
&\multicolumn{2}{c}{GLM-4} &\textbf{1.63} &\underline{2.38} &\underline{2.06} &1.00 \\
&\multicolumn{2}{c}{CPsyCounX} &\underline{1.44} &\textbf{2.81} &\textbf{2.25} &1.00 \\\midrule
\multirow{5}{*}{Social Relationship} &\multicolumn{2}{c}{InternLM2-7B-Chat} &1.31 &2.13 &1.81 &1.00 \\
&\multicolumn{2}{c}{SoulChat} &1.25 &\underline{2.31} &\textbf{2.31} &1.00 \\
&\multicolumn{2}{c}{ChatGPT} &\underline{1.38} &2.13 &2.06 &1.00 \\
&\multicolumn{2}{c}{GLM-4} &\underline{1.38} &2.19 &\underline{2.25} &1.00 \\
&\multicolumn{2}{c}{CPsyCounX} &\textbf{1.44} &\textbf{2.69} &\underline{2.25} &1.00 \\\midrule
\multirow{5}{*}{Sex} &\multicolumn{2}{c}{InternLM2-7B-Chat} &1.29 &\underline{2.14} &1.71 &1.00 \\
&\multicolumn{2}{c}{SoulChat} &1.21 &\underline{2.14} &\underline{2.21} &1.00 \\
&\multicolumn{2}{c}{ChatGPT} &\textbf{1.43} &\textbf{2.29} &\textbf{2.29} &1.00 \\
&\multicolumn{2}{c}{GLM-4} &\underline{1.36} &\underline{2.14} &2.14 &1.00 \\
&\multicolumn{2}{c}{CPsyCounX} &1.29 &\textbf{2.29} &\underline{2.21} &1.00 \\\midrule
\multirow{5}{*}{Career} &\multicolumn{2}{c}{InternLM2-7B-Chat} &\underline{1.38} &2.25 &1.19 &1.00 \\
&\multicolumn{2}{c}{SoulChat} &1.19 &2.19 &\textbf{2.25} &1.00 \\
&\multicolumn{2}{c}{ChatGPT} &1.31 &2.25 &2.00 &1.00 \\
&\multicolumn{2}{c}{GLM-4} &\textbf{1.44} &\underline{2.31} &\underline{2.19} &1.00 \\
&\multicolumn{2}{c}{CPsyCounX} &1.31 &\textbf{2.56} &2.13 &1.00 \\\midrule
\multirow{5}{*}{Mental Disease} &\multicolumn{2}{c}{InternLM2-7B-Chat} &1.25 &2.25 &1.42 &1.00 \\
&\multicolumn{2}{c}{SoulChat} &1.17 &2.17 &\underline{2.25} &1.00 \\
&\multicolumn{2}{c}{ChatGPT} &1.25 &2.25 &2.08 &1.00 \\
&\multicolumn{2}{c}{GLM-4} &\underline{1.42} &\underline{2.33} &2.17 &1.00 \\
&\multicolumn{2}{c}{CPsyCounX} &\textbf{1.50} &\textbf{2.75} &\textbf{2.33} &1.00 \\\midrule
\multirow{5}{*}{Total Average} &\multicolumn{2}{c}{InternLM2-7B-Chat} &1.30 &2.16 &1.48 &1.00 \\
&\multicolumn{2}{c}{SoulChat} &1.22 &2.18 &\underline{2.24} &1.00 \\
&\multicolumn{2}{c}{ChatGPT} &1.32 &2.25 &2.09 &1.00 \\
&\multicolumn{2}{c}{GLM-4} &\textbf{1.44} &\underline{2.36} &2.22 &1.00 \\
&\multicolumn{2}{c}{CPsyCounX} &\underline{1.39} &\textbf{2.65} &\textbf{2.29} &1.00 \\
\bottomrule[1.5pt]
\end{tabular}}
\caption{Full results of extrinsic evaluation on CPsyCounX and other baseline models. The best score for each counseling topic is \textbf{in-bold}, while the
second best score is \underline{underlined}.}
\label{tab:full_results_extrinsic_evaluation}
\end{table*}

\end{CJK*}

\end{document}